\documentclass{article}
\usepackage{ijcai26}

\usepackage{times}
\usepackage{soul}
\usepackage{url}
\usepackage[hidelinks]{hyperref}
\usepackage[utf8]{inputenc}
\usepackage[small]{caption}
\usepackage{graphicx}
\usepackage{amsmath}
\usepackage{amsthm}
\usepackage{booktabs}
\usepackage{algorithm}
\usepackage{algorithmic}
\usepackage[switch]{lineno}

\usepackage{amssymb}
\usepackage{pifont}
\usepackage[table]{xcolor}
\definecolor{rowgray}{gray}{0.92}
\newcommand{\yes}{\checkmark} 
\newcommand{\no}{\textcolor{lightgray}{\ding{55}}}

\title{COAL: Counterfactual and Observation-Enhanced Alignment Learning \\ for Discriminative Referring Multi-Object Tracking}

\author{
Shukun Jia$^{1,2}$
\and
Shiyu Hu$^3$\and
Yipei Wang$^{1,2}$\and
Ximeng Cheng$^{1,2}$\and
Yichao Cao$^4$\And
Xiaobo Lu$^{1,2,}$\footnote{Corresponding author. This arXiv version includes the supplementary material.} \\
\affiliations
$^1$School of Automation, Southeast University, Nanjing, China\\
$^2$Key Laboratory of Measurement and Control of Complex Systems of Engineering, Ministry of Education, Nanjing, China\\
$^3$School of Physical \& Mathematical Sciences, Nanyang Technological University, Singapore\\
$^4$Big Data Institute, Central South University, Changsha, China
\emails
jia\_shukun@seu.edu.cn,
xblu2013@126.com
}

\begin{document}

\maketitle

\begin{abstract}
    Referring Multi-Object Tracking (RMOT) faces a fundamental structural contradiction between the high-discriminability demand and the sparse semantic supervision. 
    This mismatch is particularly acute in highly homogeneous scenarios that require fine-grained discrimination over complex compositional semantics.
    However, under sparse supervision, models overfit to salient yet insufficient cues, thereby encouraging shortcut learning and semantic collapse.
    To resolve this, we propose \textbf{COAL} (\textbf{C}ounterfactual \& \textbf{O}bservation-enhanced \textbf{A}lignment \textbf{L}earning), a framework that advances RMOT beyond isolated structural optimization through knowledge regularization.
    First, we introduce Explicit Semantic Injection (ESI) via a VLM to densify the observation space and enhance instance discriminability. 
    Second, leveraging LLM reasoning, we propose Counterfactual Learning (CFL) to augment supervision, enforcing strict attribute verification for robust compositional recognition.
    These strategies are unified within a Hierarchical Multi-Stream Integration (HMSI) architecture, which distills external knowledge into domain-specific discriminative  representations.
    Experiments on Refer-KITTI and Refer-KITTI-V2 benchmarks validate COAL's efficacy. Notably, it surpasses the state-of-the-art by 7.28\% HOTA on the highly challenging Refer-KITTI-V2. 
    These results demonstrate the effectiveness of knowledge regularization for resolving the sparsity–discriminability paradox in RMOT.
\end{abstract}

\section{Introduction}
\label{sec:intro}

\begin{figure}[t]
\vspace{-10pt}
    \centering
    \includegraphics[width=0.9\linewidth]{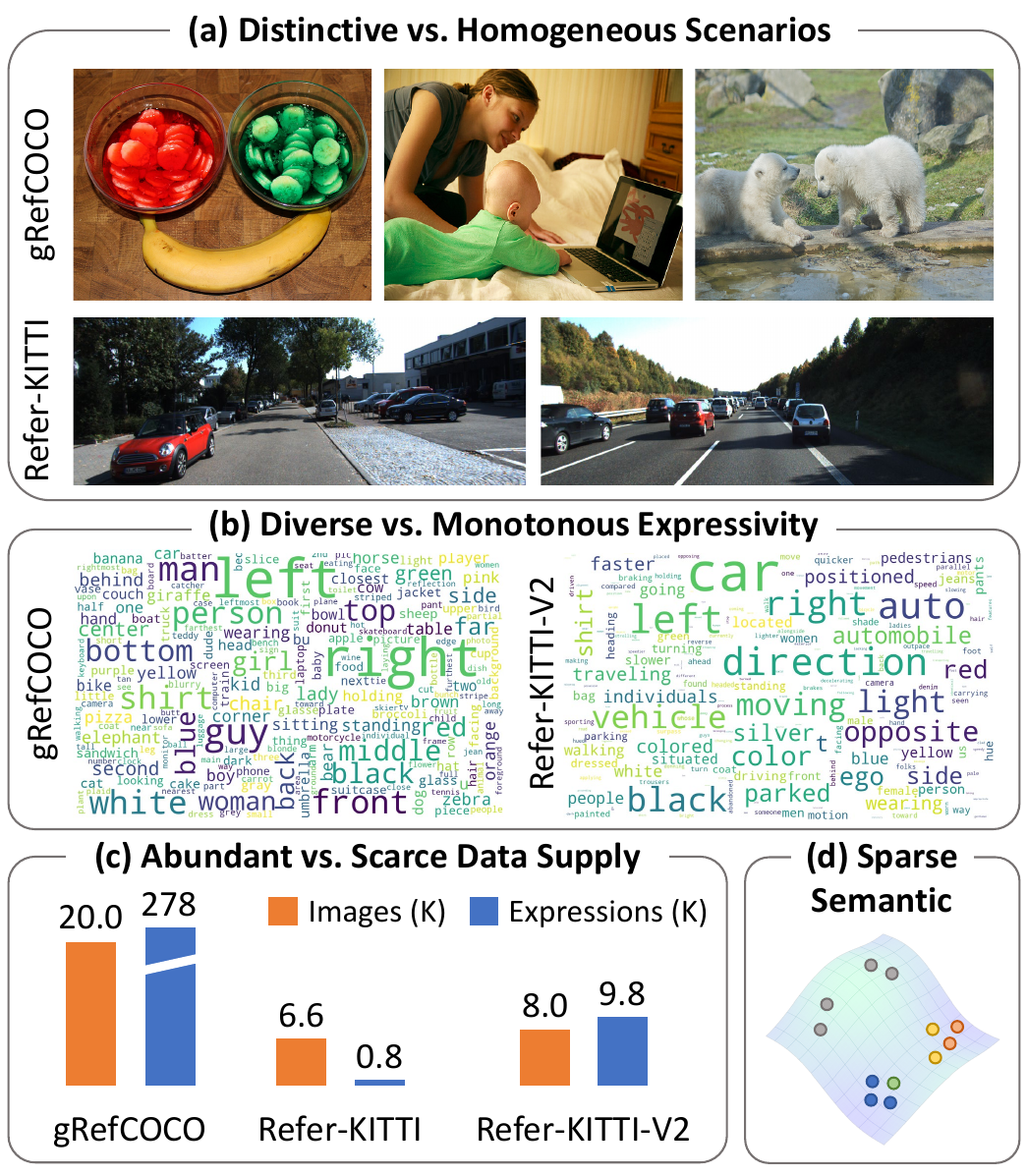}
    \vspace{-10pt}
    \caption{
    \textbf{The Sparsity-Discriminability Paradox in RMOT.} We illustrate the structural contradiction inherent to RMOT: The high-discriminability demand imposed by visual homogeneity (a) and restricted vocabulary (b) is confronted with the sparsely populated semantic manifold (d) resulting from the limited data scale (c). 
    }
    \label{fig:task_challenge}
    \vspace{-10pt}
\end{figure}

\begin{figure*}[t] 
    \centering 
    \includegraphics[width=0.9\textwidth]{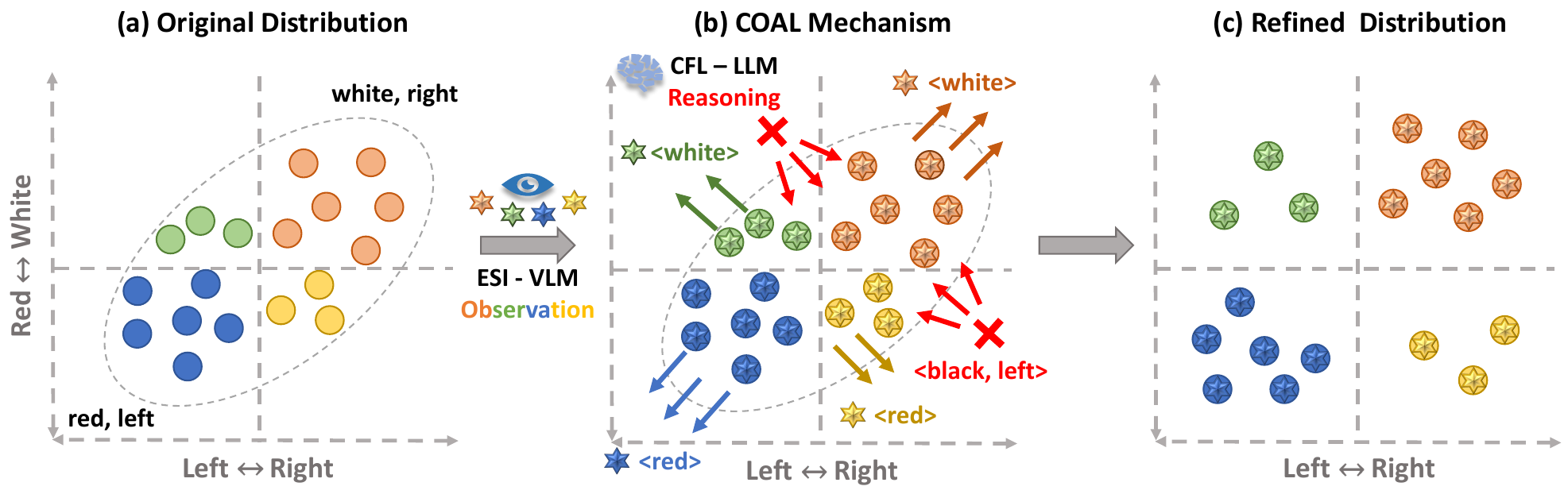} 
    \vspace{-10pt}
    \caption{\textbf{Conceptual Mechanism of COAL.} Due to sparse semantic supervision, attributes tend to be coupled in the original distribution (a). ESI injects VLM-derived discriminative cues, denoted by star symbols, to enhance semantic separability, while CFL introduces LLM-derived counterfactual supervision, denoted by red crosses, to enforce compositional discrimination (b). Consequently, the refined representation becomes semantically discriminative and structurally comprehensive (c).} 
    \label{fig:core} 
    \vspace{-10pt}
\end{figure*}

Referring Multi-Object Tracking (RMOT) requires an agent to recognize and track arbitrary objects specified by natural language descriptions. Unlike conventional Multi-Object Tracking (MOT), which primarily operates within the visual space, RMOT fundamentally demands fine-grained cross-modal discrimination.

Despite great advancements, existing methods \cite{zhuang2025cgatracker,zhao2025hff,li2025visual} predominantly formulate RMOT as a multi-modal fusion problem, implicitly assuming that the discrimination bottleneck can be resolved in isolation by increasing interaction intricacy or architectural complexity. However, this perspective overlooks a more fundamental structural contradiction inherent to the task: the mismatch between the high-discriminability demand and the extremely sparse semantic supervision. Unlike general visual grounding \cite{liu2023gres}, this contradiction is particularly pronounced in RMOT, as illustrated in Fig.~\ref{fig:task_challenge}.

First, RMOT scenarios exhibit severe visual homogeneity (Fig.~\ref{fig:task_challenge}a). In representative benchmarks like Refer-KITTI \cite{wu2023referring} and Refer-KITTI-V2 \cite{zhang2024bootstrapping}, targets share high visual and semantic similarity with numerous homogeneous distractors. This renders coarse-grained discrimination strategies ineffective, necessitating the recognition of subtle, non-salient attributes. 
Second, RMOT expressions are characterized by restricted lexical composition (Fig.~\ref{fig:task_challenge}b), as reflected by the highly repetitive attribute combinations in benchmark annotations. Lacking unique keywords or rich contextual cues, the model is forced to distinguish targets by resolving the rigorous combination of limited attributes. For example, distinguishing ``the red car on the left'' from ``the white car on the left'' or ``the red car on the right'' requires accurate compositional interpretation beyond coarse attribute matching.
Finally, these discrimination requirements face severe supervision scarcity: the limited data scale (Fig.~\ref{fig:task_challenge}c) leaves the vast combinatorial space of attributes severely undersampled, giving rise to a sparse semantic manifold (Fig.~\ref{fig:task_challenge}d).

The coexistence of fine-grained discrimination requirements and sparse supervision can induce a systematic perception bias. Models tend to prioritize easily learnable but insufficient attributes as primary discriminators, leading to attribute coupling and spurious correlations. Consequently, the model degenerates into shortcut learning, resulting in feature collapse. 
Fundamentally, this issue represents a breakdown in compositional generalization, rather than merely insufficient representation capacity or fusion architecture design.

To this end, we propose \textbf{COAL} (\textbf{C}ounterfactual \& \textbf{O}bservation-enhanced \textbf{A}lignment \textbf{L}earning), a framework designed to enrich the representation space and prevent shortcut learning by leveraging external knowledge.
First, we address semantic ambiguity via Explicit Semantic Injection (ESI). To enhance discrimination, we utilize a VLM to generate dual-purpose priors comprising visual detections and linguistic captions. To integrate these priors, we propose a Hierarchical Multi-Stream Integration (HMSI) architecture with a progressive interaction strategy: visually, pixel-word contextualization and deformable sampling transform static crops into query-aware representations; semantically, the referring query filters the caption stream to yield informative descriptions. By fusing these streams, we construct a holistic object representation that is both visually grounded and linguistically explicit.
Second, we alleviate sparse supervision via Counterfactual Learning (CFL). To prevent shortcut learning induced by limited data, we leverage an LLM to generate counterfactual hard negatives through stochastic attribute perturbation. These high-interference distractors supervise the model to disentangle subtle cues from salient ones, effectively expanding the semantic representation and mitigating feature collapse. The conceptual mechanism of COAL is illustrated in Fig.~\ref{fig:core}. Starting from the coupled original distribution, ESI enhances semantic separability through VLM-derived discriminative cues, while CFL promotes compositional discrimination through counterfactual supervision. The resulting representation is expected to be both discriminative and structurally comprehensive, which is further supported by our visualization and quantitative experiments.

\begin{figure*}[t]
    \centering
    \includegraphics[width=0.9\linewidth]{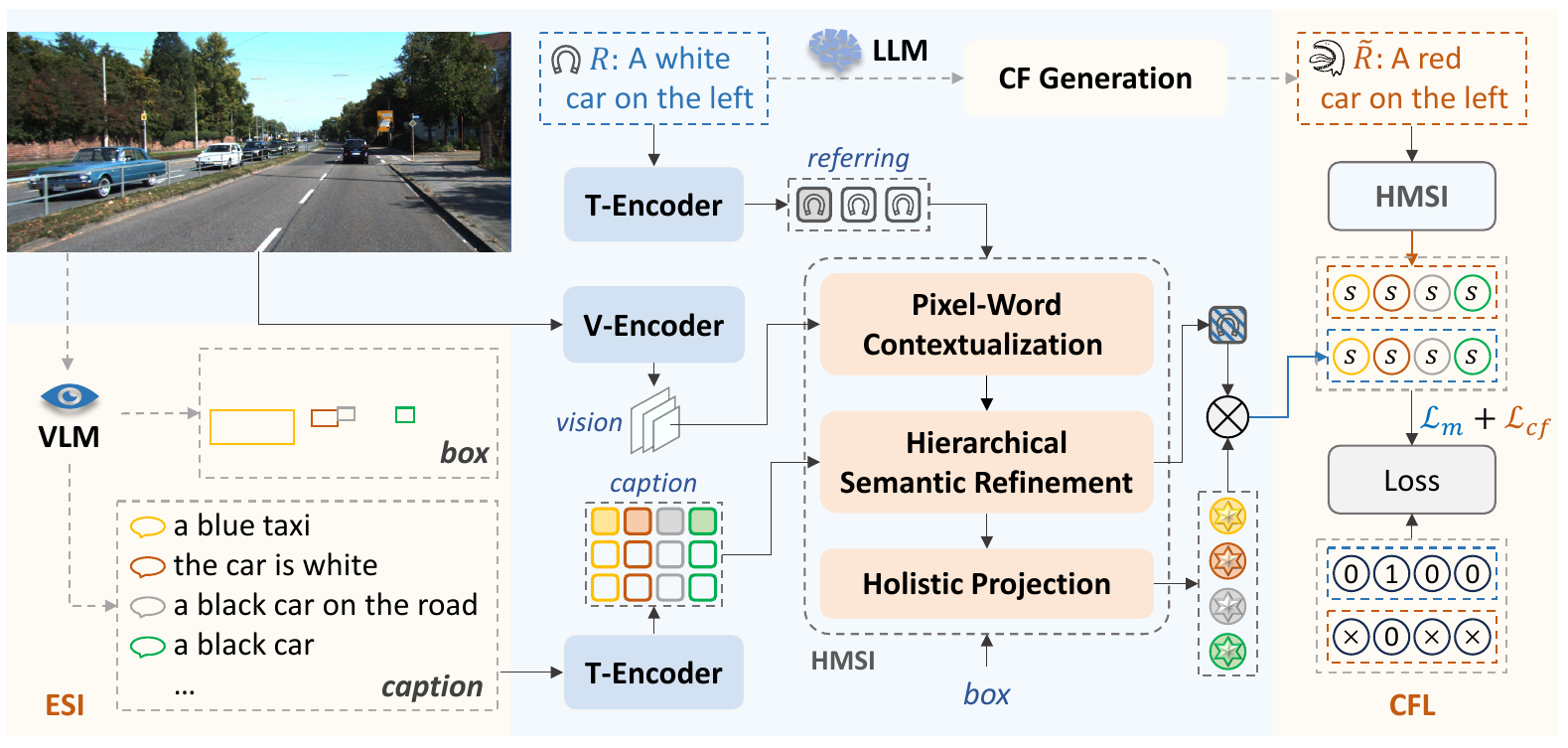}
    \vspace{-10pt}
    \caption{
    \textbf{Overview of the COAL Framework.} It leverages external knowledge through Explicit Semantic Injection (ESI) and Counterfactual Learning (CFL), unified by the Hierarchical Multi-Stream Integration (HMSI) architecture to construct semantically enriched and discriminative object representations. The V-Encoder and T-Encoder refer to the CLIP visual and textual backbones for feature extraction.
    }
    \label{fig:overall}
    \vspace{-10pt}
\end{figure*}

The contributions of this work can be summarized as:
(1) We systematically identify the Sparsity-Discriminability Paradox in RMOT, where the mismatch between fine-grained discrimination demands and sparse semantic supervision induces shortcut learning and weak compositional generalization.
(2) We propose COAL, a unified framework that leverages external knowledge to resolve this paradox. Specifically, ESI enriches semantic representations, CFL introduces counterfactual supervision, and HMSI integrates these priors for knowledge-regularized learning.
(3) COAL achieves state-of-the-art performance on both Refer-KITTI and Refer-KITTI-V2 benchmarks, surpassing the second-best method by 7.28\% HOTA on the highly challenging Refer-KITTI-V2, demonstrating the effectiveness of knowledge regularization beyond isolated structural optimization.

\section{Related Work}

\textbf{Referring Multi-Object Tracking (RMOT).} Since the establishment of the Refer-KITTI benchmark by TransRMOT \cite{wu2023referring}, RMOT research has primarily focused on isolated optimization of cross-modal interactions. To enhance feature granularity, DeepRMOT \cite{he2024visual} and HFF-Tracker \cite{zhao2025hff} introduce hierarchical fusion modules, while CGATracker \cite{zhuang2025cgatracker} and MGLT \cite{chen2025multi} utilize graph alignment and query bootstrapping to preserve linguistic cues. Other approaches like TellTrack \cite{huang2024tell} and FlexHook \cite{li2025just} refine encoder-decoder designs to address data imbalance. iKUN \cite{du2024ikun} adopts a plug-and-play strategy with a Neural Kalman Filter. Addressing temporal dynamics, TempRMOT \cite{zhang2024bootstrapping} and SKTrack \cite{li2025visual} incorporate motion modules, while works like LaMOT \cite{li2025lamot}, MLS-Track \cite{ma2024mls} and EchoTrack \cite{lin2024echotrack} expand the task scope to diverse scenarios. CDRMT \cite{liang2025cognitive} explores cognitive pathways for description disentanglement.
Similar explorations on multi-modal perception and spatio-temporal modeling have also been studied in related tracking tasks \cite{hu2024toward,hu2025adaptive,hu2025exploiting,hu2026curriculum}.
Despite significant breakthroughs, these methods remain confined to a closed data loop. In contrast, we formulate the problem from a data-centric perspective, aiming to mitigate inherent biases in limited data.

\textbf{Foundation Models in RMOT.}
Recently, Multi-modal LLMs (MLLMs) have been applied to RMOT, reflecting an emerging trend of leveraging open-world knowledge for domain-specific challenges. For instance, ReferGPT \cite{chamiti2025refergpt} employs MLLMs for zero-shot tracking via 3D-aware priors, while VMRMOT \cite{lv2025vision} utilizes them to extract motion features for enhanced alignment. In contrast to these approaches, which primarily focus on optimizing multi-modal matching or fusion, we adopt a data-centric perspective to address RMOT's structural contradiction, leveraging external knowledge to regularize the learning process and boost discriminative capacity.

\section{Methodology}

We propose COAL (Counterfactual \& Observation-enhanced Alignment Learning), a framework designed to resolve the sparsity-discriminability paradox in RMOT via external knowledge regularization. Adopting a Tracking-by-Detection \cite{zhang2022bytetrack} paradigm, COAL focuses on enhancing the semantic discriminability of observations within the tracking pipeline. As illustrated in Fig.~\ref{fig:overall}, the framework is built upon three primary components:
(1) \textbf{Explicit Semantic Injection} (\textbf{ESI}, Sec. \ref{sec:esi}): We first establish a robust observation basis via a VLM, generating dual-purpose priors (boxes and captions) to densify the sparse visual signals.
(2) \textbf{Hierarchical Multi-Stream Integration (HMSI)}: To synthesize these priors with visual and linguistic streams, we design the HMSI architecture, integrating the information flow through three progressive stages: Pixel-Word Contextualization (Sec. \ref{sec:vlc}) for early implicit grounding, Hierarchical Semantic Refinement (Sec. \ref{sec:refiner}) for explicit caption denoising, and Holistic Projection (Sec. \ref{sec:fusion}) for unifying multi-modal representations.
(3) \textbf{Counterfactual Learning} (\textbf{CFL}, Sec. \ref{sec:cflearning}): Driven by an LLM, we generate counterfactual negatives to impose a causal constraint on supervision to rectify shortcut learning.

\begin{figure}[t]
    \centering
    \includegraphics[width=1.0\linewidth]{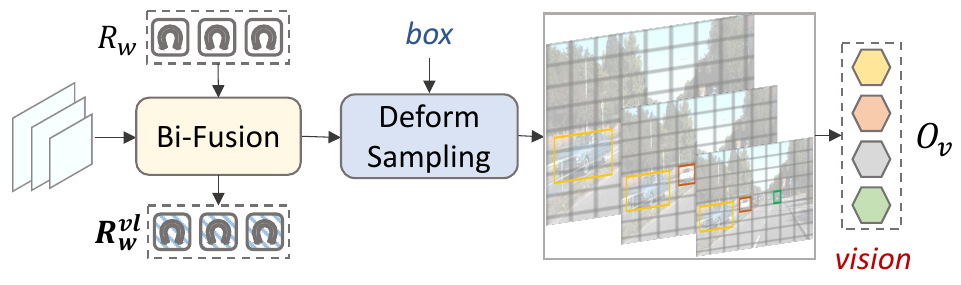}
    \vspace{-20pt}
    \caption{
    \textbf{Illustration of Pixel-Word Contextualization.}  It performs bi-directional interaction between visual and referring features, followed by VLM-guided deformable sampling to extract query-aware visual representations.
    }
    \label{fig:vl_contextualize}
    \vspace{-10pt}
\end{figure}

\subsection{Explicit Semantic Injection}
\label{sec:esi}

To mitigate the ambiguity of implicit visual features supervised by scarce data, we introduce Explicit Semantic Injection (ESI) via a frozen VLM. For each video frame, the VLM executes two parallel tasks: (1) Localization: generating dense object proposals $\mathcal{B}$; and (2) Description: producing descriptive captions $\mathcal{C}$ that explicitly materialize subtle visual cues (e.g., fine-grained color).

This mechanism establishes a dual-purpose observation prior, primarily densifying the semantic representation to serve as the foundational input for the subsequent integration network. Furthermore, by offloading detection and description to the VLM, ESI significantly alleviates the optimization burden, allowing the model to focus on fine-grained multi-modal understanding rather than struggling with basic perception tasks under sparse supervision.

\subsection{Pixel-Word Contextualization}
\label{sec:vlc}
To bridge the modality gap early in the encoding stage, we implement Pixel-Word Contextualization. Given an image and a referring sentence, we extract flattened visual features $F_v$ and tokenized word embeddings $R_w = \{w_1, \dots, w_L\}$ using frozen CLIP encoders \cite{radford2021learning}. As displayed in Fig.~\ref{fig:vl_contextualize}, instead of processing them independently, we employ Bi-Fusion \cite{liu2024grounding,chen2025multi} to perform bidirectional cross-modal interaction, allowing them to mutually exchange contextual cues:
\begin{equation}
    F_v^{vl}, R_w^{vl} = \text{Bi-Fusion}(F_v, R_w).
\end{equation}
This yields grounded visual representations $F_v^{vl}$ that highlight referred regions, and visually contextualized referring word embeddings $R_w^{vl}$ that drive the subsequent refinement.

\textbf{Visual Stream.} To extract object-centric features, we utilize the high-quality proposals $\mathcal{B}=\{b_i\}_{i=1}^N$ generated by the VLM. Guided by these boxes, we reconstruct the grounded representations $F_v^{vl}$ into a multi-scale feature map $\mathrm{MS}(F_v^{vl})$ and apply the standard deformable sampling \cite{zhu2020deformable} to dynamically aggregate object-centric features around proposal regions:
\begin{equation}
    O_{v, i} = \text{DeformSample}(\mathrm{MS}(F_v^{vl}), \ b_i).
\end{equation}
Sampled on the linguistically contextualized feature map, $O_v$ encodes both geometric details and linguistic relevance.

\begin{figure}[t]
    \centering
    \includegraphics[width=0.8\linewidth]{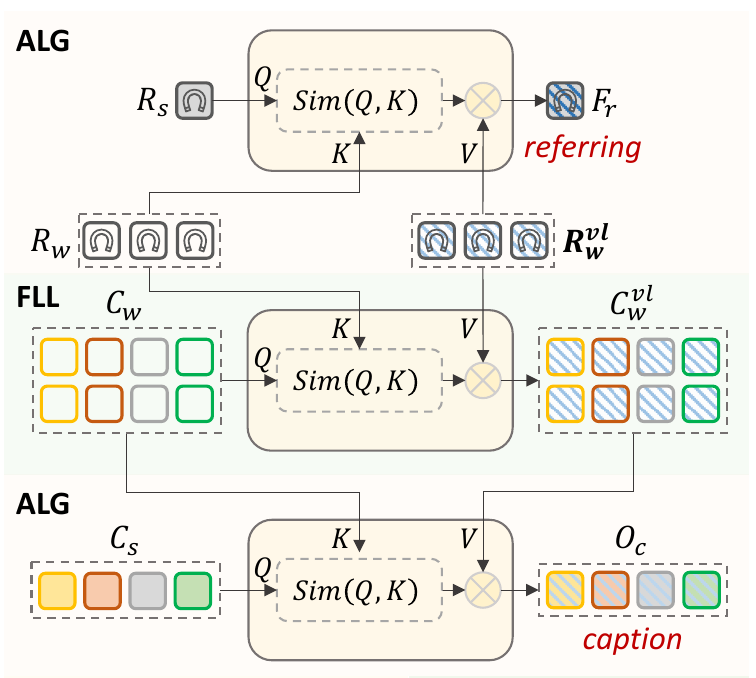}
    \vspace{-10pt}
    \caption{
    \textbf{Illustration of Hierarchical Semantic Refinement.} It refines referring and caption streams via a hierarchical mechanism: Filtering from Local to Local (FLL), to denoise captions using referring cues, and Aggregation from Local to Global (ALG), to synthesize holistic semantic descriptors.
    }
    \label{fig:hire_module}
    \vspace{-10pt}
\end{figure}

\subsection{Hierarchical Semantic Refinement}
\label{sec:refiner}

Parallel to the visual stream, we propose this module to refine both referring and caption streams (Fig.~\ref{fig:hire_module}) through hierarchical filtering and aggregation, ensuring contextually aligned and semantically denoised linguistic representations.

\textbf{Referring Stream.} We synthesize local linguistic cues through the ALG (Aggregation from Local to Global) mechanism. Specifically, the visually-enhanced word features $R_w^{vl}$ are aggregated via cross-attention using the sentence embedding $R_s$ as the query, yielding a referring representation $F_r$ enriched with visual context:
\begin{equation}
    F_r = \text{CrossAttn}(Q=R_s, K=R_w, V=R_w^{vl}).
\end{equation}

\textbf{Caption Stream.} 
We treat the descriptive captions $\mathcal{C}$ as an explicit semantic stream, tokenized into word and sentence embeddings $C_w$, $C_s$, respectively. To filter redundant information, we first propose the FLL (Filtering from Local to Local) mechanism, leveraging the textual alignment between caption and referring words to inject visual context $R_w^{vl}$:
\begin{equation}
    C_w^{vl} = \text{CrossAttn}(Q=C_w, K=R_w, V=R_w^{vl}).
\end{equation}
This operation activates caption tokens that resonate with the referring query while enriching them with contextual visual cues. 
Subsequently, we employ the ALG to synthesize the filtered caption words $C_w^{vl}$ using the sentence embedding $C_s$:
\begin{equation}
    O_c = \text{CrossAttn}(Q=C_s, K=C_w, V=C_w^{vl}).
\end{equation}
The resulting explicit semantic feature $O_c$ serves as a query-adaptive descriptor, ready for downstream holistic fusion.

\subsection{Holistic Projection}
\label{sec:fusion}

To construct a comprehensive representation, we perform a joint projection of the implicit visual stream ($O_v$), explicit semantic stream ($O_c$), and spatial geometry. Box coordinates of $\mathcal{B}$ are first encoded into sinusoidal positional embeddings $O_p$. Treating captions as the semantic counterpart to vision, we construct spatially-aware representations for both streams:
\begin{equation}
\label{eq:two_modal}
    \tilde{O}_v = O_v \oplus O_p, \quad \tilde{O}_c = O_c \oplus O_p,
\end{equation}
where $\tilde{O}_v$ and $\tilde{O}_c$ encapsulate the object's identity within the implicit visual and explicit knowledge spaces, respectively. $\oplus$ represents element-wise addition. To synthesize these complementary views, we project them into a unified latent space:
\begin{equation}
\label{eq:holistic}
    F_o = \mathcal{F}_{\text{fuse}}(\tilde{O}_v \oplus \tilde{O}_c),
\end{equation}
where $\mathcal{F}_{\text{fuse}}$ denotes a linear projection. This yields a holistic object query $F_o$ that is visually grounded, linguistically explicit, and spatially precise. The final matching score is computed as the cosine similarity between the object query and the referring query: $S = \cos(F_o, F_r)$.

\subsection{Counterfactual Learning}
\label{sec:cflearning}
To address attribute coupling inherently under semantic sparsity, we introduce Counterfactual Learning (CFL). As illustrated in the right panel of Fig.~\ref{fig:overall}, CFL imposes a causal constraint on supervision to enforce attribute disentanglement.

\textbf{Counterfactual Generation.} For a referring expression $R$, we harness the reasoning capabilities of an LLM to generate a hard negative query $\tilde{R}$ via stochastic perturbation. The LLM parses attributes and randomly replaces one (e.g., altering Color while retaining the others).
This stochastic strategy functions as an implicit hard negative miner: perturbing a dominant attribute creates an easy negative with minimal loss, whereas perturbing a non-dominant attribute creates a hard negative with high similarity to the target, thus dominating the gradient. This naturally drives the model to focus on discerning subtle, fine-grained attributes beyond salient ones.

\textbf{Optimization Objective.} We formulate the training objective using Binary Cross Entropy (BCE). To stabilize gradient propagation, we linearly map the cosine similarity score $S(u, v) \in [-1, 1]$ to a probability space:
\begin{equation}
    P(u, v) = \frac{1}{2} (S(u, v) + 1).
\end{equation}
The total loss $\mathcal{L} = \mathcal{L}_{m} + \mathcal{L}_{cf}$ comprises two complementary terms.
First, the main grounding loss ($\mathcal{L}_{m}$) is computed over all $N$ detected objects to minimize the classification error:
\begin{equation}
    \mathcal{L}_{m} = \frac{1}{N} \sum_{i=1}^{N} \ell_i, \quad 
    \ell_i = \begin{cases} 
        -\log(p_i) & \text{if } y_i = 1, \\
        -\log(1 - p_i) & \text{if } y_i = 0,
    \end{cases}
\end{equation}
where $y_i$ is the ground-truth label, and $p_i=P(F_{o,i}, F_r)$ denotes the matching probability between the $i$-th object and the global referring query. This term ensures the model correctly identifies the target while suppressing non-target objects.

Second, the counterfactual loss ($\mathcal{L}_{cf}$) is applied exclusively to the target object, which is denoted as $o^+$. Crucially, since the HMSI network is query-guided, the target's holistic representation dynamically updates from $F_{o^+}$ to $F_{o^+}^{cf}$ when conditioned on the counterfactual query $F_r^{cf}$.
To enforce attribute disentanglement, we require the model to suppress the target under the perturbed description:
\begin{equation}
    \mathcal{L}_{cf} = - \log(1 - P(F_{o^+}^{cf}, \ F_r^{cf})).
\end{equation}
Note that we only penalize the original target object $o^+$ while ignoring other objects, as the perturbed query $F_r^{cf}$ might coincidentally match another object in the scene.

By minimizing the joint objective, our framework establishes a dual-force optimization mechanism. $\mathcal{L}_{m}$ exerts a \textit{Pull} force for standard alignment. In contrast, $\mathcal{L}_{cf}$ exerts a \textit{Push} force: by teaching the model \textit{what the object is not}, it penalizes spurious correlations against subtle perturbations. This interplay compels the model to acquire compositional understanding, effectively expanding the semantic representation to prevent feature collapse induced by shortcut learning.

\section{Experiments}

\subsection{Datasets and Evaluation Metrics}
We systematically evaluate our method on Refer-KITTI \cite{wu2023referring} and Refer-KITTI-V2 \cite{zhang2024bootstrapping} to validate its effectiveness under diverse semantic and discriminative conditions. As the most representative and widely adopted benchmarks in RMOT, these two datasets collectively span a comprehensive difficulty spectrum, extending from fundamental scenarios to rigorous high-discriminability challenges.
Refer-KITTI established the initial benchmark, comprising 15 training and 3 testing videos with 818 expressions.
Refer-KITTI-V2 is a large-scale extension featuring 9,758 diverse expressions across 21 videos. This increased complexity imposes a stringent test on fine-grained discrimination, making it a critical benchmark for validating our framework. We perform ablation studies on both datasets to verify the consistent contribution of the proposed modules.

For evaluation, we adopt HOTA \cite{luiten2021hota} as the primary metric to balance detection and association. We also report DetA and AssA for detailed insights. Moreover, we include auxiliary metrics: DetRe/DetPr to characterize detection performance, AssRe/AssPr to quantify identity consistency, and LocA to assess bounding box quality.

\subsection{Implementation Details}

We provide comprehensive implementation details, including data preprocessing, network architecture, training strategy, and inference procedure, in the supplementary material. The VLM and LLM components are used offline for proposal/caption generation and counterfactual preparation, respectively, introducing no additional online reasoning overhead during inference.

\begin{table*}[t]
    \footnotesize
    \centering
    \begin{tabular}{c c c c c c c c c c}
        \toprule
        \textbf{Method} & \textbf{Publication} & \textbf{HOTA}$\uparrow$ & \textbf{DetA}$\uparrow$ & \textbf{AssA}$\uparrow$ & \textbf{DetRe}$\uparrow$ & \textbf{DetPr}$\uparrow$ & \textbf{AssRe}$\uparrow$ & \textbf{AssPr}$\uparrow$ & \textbf{LocA}$\uparrow$ \\
        \midrule
        \multicolumn{10}{c}{\textbf{\textit{Refer-KITTI}}} \\
        TransRMOT \cite{wu2023referring}    & CVPR 2023 & 46.56 & 37.97 & 57.33 & 49.69 & 60.10 & 60.02 & 89.67 & 90.33 \\
        DeepRMOT \cite{he2024visual}     & ICASSP 2024 & 39.55 & 30.12 & 53.23 & 41.91 & 47.47 & 58.47 & 82.16 & 80.49 \\
        iKUN \cite{du2024ikun}         & CVPR 2024 & 48.84 & 35.74 & 66.80 & 51.97 & 52.25 & 72.95 & 87.09 & -- \\
        EchoTrack \cite{lin2024echotrack}    & ITS 2024 & 48.86 & 41.26 & 57.59 & 53.42 & 62.83 & 61.61 & 89.33 & 90.74 \\
        TempRMOT \cite{zhang2024bootstrapping}    & arXiv 2024 & 52.21 & 40.95 & 66.75 & 55.65 & 59.25 & 71.82 & 87.76 & 90.40 \\
        CGATracker \cite{zhuang2025cgatracker}  & TCSVT 2025 & 48.81 & 38.66 & 61.77 & 51.67 & 59.12 & 65.87 & \textbf{90.26} & -- \\
        MGLT  \cite{chen2025multi}       & TIM 2025 & 49.25 & 37.09 & 65.50 & -- & -- & -- & -- & -- \\
        CDRMT \cite{liang2025cognitive}       & IF 2025    & 49.35 & 40.34 & 60.56 & 54.54 & 59.30 & 64.70 & 89.80 & 90.61\\
        SKTrack \cite{li2025visual}     & TMM 2025  & 50.85 & \textbf{42.24} & 61.39 & 58.10 & 59.24 & 66.68 & 88.76 & 90.43 \\
        HFF-Tracker \cite{zhao2025hff}  & AAAI 2025  & 52.41 & 41.29 & 66.65 & 53.42 & \textbf{62.89} & 71.48 & 88.96 & 90.76 \\
        \rowcolor{rowgray}
        \textbf{COAL} & \textbf{Ours} & \textbf{53.38} & 40.89 & \textbf{69.74} & \textbf{60.32} & 54.73 & \textbf{76.25} &  86.68 & \textbf{90.99} \\

        \midrule[\heavyrulewidth]
        \multicolumn{10}{c}{\textbf{\textit{Refer-KITTI-V2}}} \\
        TransRMOT \cite{wu2023referring}   & CVPR 2023 & 31.00 & 19.40 & 49.68 & 36.41 & 28.97 & 54.59 & 82.29 & 89.82    \\
        iKUN \cite{du2024ikun}        & CVPR 2024 & 10.32 & 2.17 & 49.77 & 2.36 & 19.75 & 58.48 & 68.64 & 74.56   \\
        TempRMOT \cite{zhang2024bootstrapping}    & arXiv 2024 & 35.04 & 22.97 & 53.58 & 34.23 & 40.41 & 59.50 & 81.29 & 90.07 \\
        CDRMT  \cite{liang2025cognitive}      & IF 2025 & 31.99 & 20.37 & 50.35 & 26.40 & \textbf{46.26} & 53.40 & \textbf{85.90} & 90.36 \\
        CGATracker \cite{zhuang2025cgatracker}  & TCSVT 2025 & 33.19 & 22.04 & 50.13 & 38.00 & 33.88 & 54.73 & 84.90 & -- \\
        SKTrack \cite{li2025visual}     & TMM 2025 & 35.29 & 23.87 & 52.35 & 39.97 & 36.48 &  57.45 & 84.23 & 88.89 \\
        HFF-Tracker \cite{zhao2025hff}  & AAAI 2025  & 36.18 & 24.64 & 53.27 & 36.86 & 41.83 & 59.42 & 81.40 & 89.77 \\
        \rowcolor{rowgray}
        \textbf{COAL} & \textbf{Ours} & \textbf{43.46} & \textbf{32.83} & \textbf{57.62} & \textbf{60.20} & 41.19 & \textbf{66.23} & 80.33 & \textbf{90.38} \\

        \bottomrule
    \end{tabular}
    \vspace{-5pt}
    \caption{\textbf{Comparison with State-of-the-Art Methods on the Refer-KITTI and Refer-KITTI-V2 Benchmarks.} The best results for each metric are highlighted in \textbf{bold}.}
    \label{tab:benchmark_results}
\end{table*}

\begin{table*}[t]
    \footnotesize
    \centering
    \setlength{\tabcolsep}{3.5mm} 
    \begin{tabular}{cccccccc}
        \toprule
        \multicolumn{2}{c}{\textbf{Knowledge Priors}} &
        \multicolumn{3}{c}{\textbf{Refer-KITTI}} &
        \multicolumn{3}{c}{\textbf{Refer-KITTI-V2}} \\
        \cmidrule(lr){1-2}
        \cmidrule(lr){3-5}
        \cmidrule(lr){6-8}

        \textbf{ESI} & \textbf{CFL} &
        \textbf{HOTA}$\uparrow$ &
        \textbf{DetA}$\uparrow$ &
        \textbf{AssA}$\uparrow$ &
        \textbf{HOTA}$\uparrow$ &
        \textbf{DetA}$\uparrow$ &
        \textbf{AssA}$\uparrow$ \\
        \midrule
        \no & \no & 46.84 & 32.87 & 66.76 & 37.29 & 25.12 & 55.44 \\
        
        \yes & \no & 48.64 & 35.36 & 66.95 & 39.38 & 27.66 & 56.15 \\
        
        \no & \yes & 51.48 & 39.15 & 67.75 & 40.01 & 28.43 & 56.40 \\
        
        \yes & \yes & 53.38 & 40.89 & 69.74 & 43.46 & 32.83 & 57.62 \\
        \bottomrule
    \end{tabular}
    \vspace{-5pt}
    \caption{\textbf{Ablation Study on the Impact of External Knowledge Integration.} We validate the contribution of Explicit Semantic Injection (ESI) and Counterfactual Learning (CFL) on both Refer-KITTI and Refer-KITTI-V2 benchmarks.}
    \label{tab:ablation_what}
    \vspace{-10pt}
\end{table*}

\subsection{Benchmark Results}
Tab. \ref{tab:benchmark_results} presents the comparison with state-of-the-art methods. Our COAL framework achieves the best performance across both benchmarks, validating the effectiveness of integrating external knowledge priors under semantic sparsity.

On the Refer-KITTI benchmark, COAL achieves 53.38\% HOTA, outperforming the previous state-of-the-art HFF-Tracker \cite{zhao2025hff} by 0.97\%. On the more challenging Refer-KITTI-V2, which features significantly complex semantics and higher discriminability demands, our method demonstrates a substantial advantage. COAL attains 43.46\% HOTA, surpassing the second-best method by a margin of 7.28\%.
The widening performance gap between the two datasets strongly supports our core motivation. Existing methods, focusing on internal structural optimization, tend to saturate on the relatively simpler Refer-KITTI dataset but struggle to generalize to the complex, compositional attributes in Refer-KITTI-V2. We attribute this robustness to the effective utilization of external knowledge through three key components: (1) Explicit knowledge injection (via VLM) densifies the observation space, enabling the recognition of subtle attributes that implicit features often miss; (2) Counterfactual alignment (via LLM) enforces attribute disentanglement, preventing shortcut learning in sparse semantic scenarios; and (3) Hierarchical integration (via HMSI) effectively synthesizes these diverse priors into a coherent representation. This demonstrates that knowledge-regularized integration is a highly effective strategy for resolving the sparsity-discriminability paradox, providing a promising direction beyond the pure closed-loop structural optimization for RMOT.

\subsection{Ablation Studies}

We conduct extensive ablation studies on both Refer-KITTI and Refer-KITTI-V2 benchmarks to dissect the effectiveness of our framework. The analysis is structured around two fundamental questions. (1) \textit{What to Inject:} The necessity of the introduced knowledge priors; and (2) \textit{How to inject:} The rationality of the proposed hierarchical integration architecture.

\textbf{Impact of External Knowledge Integration.} 
To investigate the contributions of the two external knowledge integration strategies, we define two ablation settings: (1) w/o ESI: We remove the Explicit Semantic Injection (ESI), which entails disabling the filtering and aggregation of captions in the Hierarchical Multi-Stream Integration (HMSI) module, relying solely on the implicit visual stream. (2) w/o CFL: We deactivate the Counterfactual Learning (CFL) branch and remove the related loss $\mathcal{L}_{cf}$ during training.
As shown in Tab. \ref{tab:ablation_what}, 
ESI consistently improves performance over the baseline. For instance, on the more challenging Refer-KITTI-V2 dataset, HOTA improves from 37.29\% to 39.38\%. This confirms that VLM captions serve as effective semantic anchors to mitigate visual ambiguity.
Introducing CFL yields even more significant gains, boosting HOTA to 40.01\%. This demonstrates its crucial role in preventing shortcut learning.
Ultimately, the full model achieves the highest performance, validating that representation enrichment (via ESI) and supervision augmentation (via CFL) operate as complementary mechanisms to address the structural sparsity-discriminability paradox in RMOT.

\begin{table*}[t]
    \footnotesize
    \centering
    \setlength{\tabcolsep}{3.5mm} 
    \begin{tabular}{cccccccc}
        \toprule
        \multicolumn{2}{c}{\textbf{Fusion Architecture}} &
        \multicolumn{3}{c}{\textbf{Refer-KITTI}} &
        \multicolumn{3}{c}{\textbf{Refer-KITTI-V2}} \\
        \cmidrule(lr){1-2}
        \cmidrule(lr){3-5}
        \cmidrule(lr){6-8}

        \textbf{Bi-Fusion} & \textbf{Caption Refine} &
        \textbf{HOTA}$\uparrow$ &
        \textbf{DetA}$\uparrow$ &
        \textbf{AssA}$\uparrow$ &
        \textbf{HOTA}$\uparrow$ &
        \textbf{DetA}$\uparrow$ &
        \textbf{AssA}$\uparrow$ \\
        \midrule
        \no & \no & 49.34 & 35.27 & 69.07 & 38.43 & 27.16 & 54.47 \\
        
        \yes & \no & 52.39 & 39.88 & 68.87 & 41.42 & 30.81 & 55.76 \\
        
        \no & \yes & 52.85 & 41.04 & 68.10 & 42.85 & 32.74 & 56.19 \\
        
        \yes & \yes & 53.38 & 40.89 & 69.74 & 43.46 & 32.83 & 57.62 \\
        \bottomrule
    \end{tabular}
    \vspace{-7pt}
    \caption{\textbf{Ablation Study on Hierarchical Multi-Stream Integration (HMSI).} We analyze the necessity of key interaction mechanisms: Bi-Fusion (the core of Pixel-Word Contextualization, Sec. \ref{sec:vlc}) and Caption Refinement (the core of Hierarchical Semantic Refinement, Sec. \ref{sec:refiner}), on both Refer-KITTI and Refer-KITTI-V2 benchmarks.}
    \label{tab:ablation_how}
    \vspace{-10pt}
\end{table*}

\textbf{Efficacy of Fusion Architecture.}
To validate our hierarchical interaction design, we compare the full HMSI architecture against simplified variants in Tab. \ref{tab:ablation_how}.
First, we investigate the impact of early pixel-word interaction by removing the Bi-Fusion module (Sec. \ref{sec:vlc}). The significant performance drop from 41.42\% to 38.43\% HOTA on Refer-KITTI-V2 confirms that the mutual injection is critical for subsequent refinement: it not only generates query-aware visual representations but also enriches referring signals with visual context.
Second, we examine the necessity of caption refinement (Sec. \ref{sec:refiner}). Substituting the referring-guided refinement $O_c$ with raw sentence-level captions $C_s$ leads to suboptimal results. The superiority of the full model validates that using the referring query as a semantic filter to denoise the caption stream is essential for extracting discriminative cues.
Interestingly, we observe a compensation effect between the two modules: The performance gain of Bi-Fusion narrows from 2.99\% to 0.61\% when caption refinement is active. This suggests that the refined caption stream can partially compensate for the lack of early cross-modal contextualization by providing robust external priors. Nevertheless, optimal performance still requires the synergy of both components.

Together, these experiments demonstrate that effective knowledge regularization requires both informative priors (\textit{what}) and hierarchical integration (\textit{how}).

\subsection{Visualization}

We visualize t-SNE embeddings of object representations from a validation sequence to qualitatively elucidate how Explicit Semantic Injection (ESI) and Counterfactual Learning (CFL) reshape the semantic manifold. By mapping the ablation settings from Tab. \ref{tab:ablation_what} to visual clusters, this analysis provides an intuitive understanding of the quantitative gains.

\begin{figure}[h]
    \centering
    \includegraphics[width=0.9\linewidth]{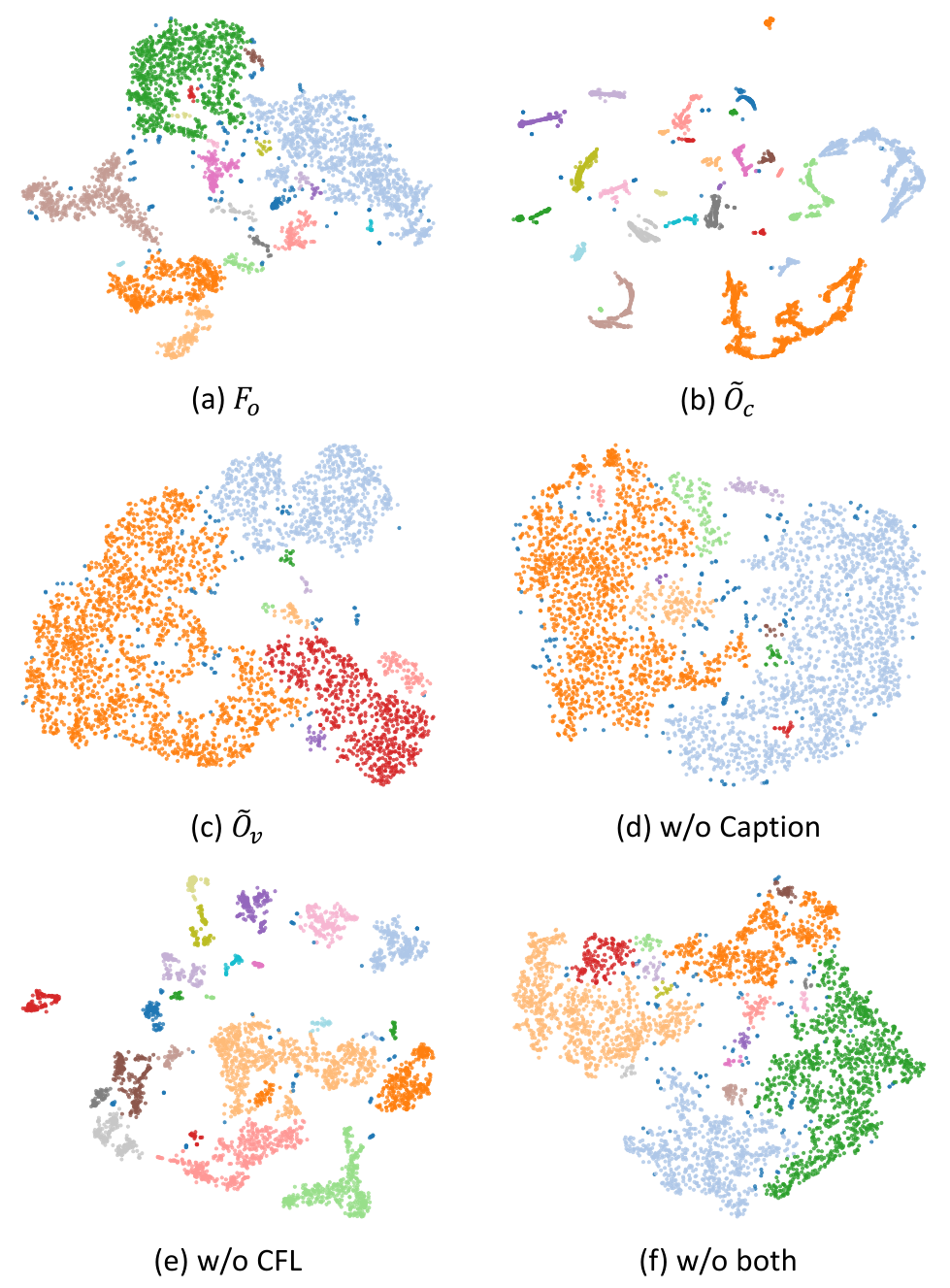}
    \vspace{-10pt}
    \caption{
    \textbf{Visualization of t-SNE Embeddings for Object Representations.} Subplots (a)-(c) illustrate feature from our full model: (a) the holistic representation $F_o$ (Equation \ref{eq:holistic}), (b) the explicit caption component $\widetilde{O}_c$ (Equation \ref{eq:two_modal}), and (c) the implicit visual component $\widetilde{O}_v$. (d)-(f) compare holistic representations from ablation variants: (d) w/o Caption, (e) w/o CFL, and (f) Baseline (w/o both). Colors indicate different semantic clusters.
    }
    \label{fig:tsne}
    \vspace{-10pt}
\end{figure}

\textbf{The Anchoring Effect of ESI.} Comparing the implicit visual stream (Fig.~\ref{fig:tsne}c) with the explicit caption stream (Fig.~\ref{fig:tsne}b) reveals a stark contrast: while visual features overlap severely due to homogeneity, captions form compact clusters, serving as highly discriminative semantic anchors. Integrating these anchors (Fig.~\ref{fig:tsne}a) effectively alleviates the ambiguity present in the baseline (Fig.~\ref{fig:tsne}f). This refinement persists even without CFL (comparing Fig.~\ref{fig:tsne}e vs. \ref{fig:tsne}f). Moreover, we observe a phenomenon of implicit knowledge distillation: the visual component of our full model (Fig.~\ref{fig:tsne}c) shows better separability than the caption-free model (Fig.~\ref{fig:tsne}d). This indicates that the caption stream acts as a semantic reference, guiding the model to discern discriminative visual cues even within the implicit branch.

\textbf{The Disentangling Effect of CFL.} Comparing the model without CFL (Fig.~\ref{fig:tsne}e) to the full model (Fig.~\ref{fig:tsne}a) elucidates the mechanism for mitigating semantic collapse. The former forms extremely tight, point-like clusters. Although superficially appealing, this is symptomatic of feature collapse by shortcut learning. In contrast, introducing CFL forces these clusters to expand and fill the semantic space (Fig.~\ref{fig:tsne}a). This dispersion is theoretically desirable: by distinguishing counterfactual hard negatives, CFL prevents the representation from degenerating into a single-attribute dominated embedding. Instead, it enforces the encoding of fine-grained nuances, resulting in a disentangled representation that faithfully reflects the intrinsic diversity of the data manifold.

In essence, the two modules operate in a complementary manner: ESI acts as an anchor, injecting distinct semantics into ambiguous visual features to sharpen semantic boundaries; meanwhile, CFL promotes expansion within these clusters, preventing feature collapse into single-dimensional shortcuts. This synergy ensures a representation that is both semantically discriminative and structurally comprehensive.

\section{Conclusion}

In this work, we identify and analyze the Sparsity--Discriminability Paradox in RMOT, where fine-grained discrimination requirements conflict with sparse semantic supervision. To address this challenge, we propose COAL, a knowledge-centric alignment framework that incorporates external semantic regularization beyond isolated structural optimization. By integrating Explicit Semantic Injection (ESI) and Counterfactual Learning (CFL) within a Hierarchical Multi-Stream Integration (HMSI) architecture, COAL alleviates shortcut learning and improves compositional discrimination under scarce supervision. Extensive experiments on Refer-KITTI and Refer-KITTI-V2 demonstrate the effectiveness of systematically introducing external semantic knowledge from foundation models for mitigating discrimination bottlenecks in small-scale RMOT tasks.

\section*{Acknowledgments}

We are grateful for the support of the National Natural Science Foun-dation of China (No. 62271143), the Frontier Technologies R\&D Program of Jiangsu (No. BF2024060), and the Big Data Computing Center of Southeast University. We thank the IDEA Research Team for providing academic access to the DINO-X API. 

\bibliographystyle{named}
\bibliography{ijcai26}

@article{ren2024dino,
  title={Dino-x: A unified vision model for open-world object detection and understanding},
  author={Ren, Tianhe and Chen, Yihao and Jiang, Qing and Zeng, Zhaoyang and Xiong, Yuda and Liu, Wenlong and Ma, Zhengyu and Shen, Junyi and Gao, Yuan and Jiang, Xiaoke and others},
  journal={arXiv preprint arXiv:2411.14347},
  year={2024}
}

@article{yang2025qwen3,
  title={Qwen3 technical report},
  author={Yang, An and Li, Anfeng and Yang, Baosong and Zhang, Beichen and Hui, Binyuan and Zheng, Bo and Yu, Bowen and Gao, Chang and Huang, Chengen and Lv, Chenxu and others},
  journal={arXiv preprint arXiv:2505.09388},
  year={2025}
}

@inproceedings{wu2023referring,
  title={Referring multi-object tracking},
  author={Wu, Dongming and Han, Wencheng and Wang, Tiancai and Dong, Xingping and Zhang, Xiangyu and Shen, Jianbing},
  booktitle={Proceedings of the IEEE/CVF conference on computer vision and pattern recognition},
  pages={14633--14642},
  year={2023}
}

@article{zhang2024bootstrapping,
  title={Bootstrapping referring multi-object tracking},
  author={Zhang, Yani and Wu, Dongming and Han, Wencheng and Dong, Xingping},
  journal={arXiv preprint arXiv:2406.05039},
  year={2024}
}

@article{ma2024mls,
  title={Mls-track: Multilevel semantic interaction in rmot},
  author={Ma, Zeliang and Yang, Song and Cui, Zhe and Zhao, Zhicheng and Su, Fei and Liu, Delong and Wang, Jingyu},
  journal={arXiv preprint arXiv:2404.12031},
  year={2024}
}

@inproceedings{du2024ikun,
  title={ikun: Speak to trackers without retraining},
  author={Du, Yunhao and Lei, Cheng and Zhao, Zhicheng and Su, Fei},
  booktitle={Proceedings of the IEEE/CVF Conference on Computer Vision and Pattern Recognition},
  pages={19135--19144},
  year={2024}
}

@article{lin2024echotrack,
  title={EchoTrack: Auditory referring multi-object tracking for autonomous driving},
  author={Lin, Jiacheng and Chen, Jiajun and Peng, Kunyu and He, Xuan and Li, Zhiyong and Stiefelhagen, Rainer and Yang, Kailun},
  journal={IEEE Transactions on Intelligent Transportation Systems},
  year={2024},
  publisher={IEEE}
}

@article{zhuang2025cgatracker,
  title={CGATracker: Correlation-Aware Graph Alignment for Referring Multi-Object Tracking},
  author={Zhuang, Siping and Li, Guangyao and Wu, Qiangqiang and Lu, Yang and Hu, Hai-Miao and Wang, Hanzi},
  journal={IEEE Transactions on Circuits and Systems for Video Technology},
  year={2025},
  publisher={IEEE}
}

@article{chen2025multi,
  title={Multi-granularity localization transformer with collaborative understanding for referring multi-object tracking},
  author={Chen, Jiajun and Lin, Jiacheng and Zhong, Guojin and Yao, You and Li, Zhiyong},
  journal={IEEE Transactions on Instrumentation and Measurement},
  year={2025},
  publisher={IEEE}
}

@article{liang2025cognitive,
  title={Cognitive Disentanglement for Referring Multi-Object Tracking},
  author={Liang, Shaofeng and Guan, Runwei and Lian, Wangwang and Liu, Daizong and Sun, Xiaolou and Wu, Dongming and Yue, Yutao and Ding, Weiping and Xiong, Hui},
  journal={Information Fusion},
  pages={103349},
  year={2025},
  publisher={Elsevier}
}

@inproceedings{chamiti2025refergpt,
  title={ReferGPT: Towards Zero-Shot Referring Multi-Object Tracking},
  author={Chamiti, Tzoulio and Di Bella, Leandro and Munteanu, Adrian and Deligiannis, Nikos},
  booktitle={Proceedings of the Computer Vision and Pattern Recognition Conference},
  pages={3849--3858},
  year={2025}
}

@article{li2025visual,
  title={Visual-Linguistic Feature Alignment with Semantic and Kinematic Guidance for Referring Multi-Object Tracking},
  author={Li, Yizhe and Zhou, Sanping and Qin, Zheng and Wang, Le},
  journal={IEEE Transactions on Multimedia},
  year={2025},
  publisher={IEEE}
}

@inproceedings{zhao2025hff,
  title={HFF-Tracker: A Hierarchical Fine-grained Fusion Tracker for Referring Multi-Object Tracking},
  author={Zhao, Zeyong and Hao, Yanchao and Zhang, Minghao and Liu, Qingbin and Li, Bo and Sui, Dianbo and He, Shizhu and Chen, Xi},
  booktitle={Proceedings of the AAAI Conference on Artificial Intelligence},
  volume={39},
  number={10},
  pages={10528--10536},
  year={2025}
}

@inproceedings{he2024visual,
  title={Visual-linguistic representation learning with deep cross-modality fusion for referring multi-object tracking},
  author={He, Wenyan and Jian, Yajun and Lu, Yang and Wang, Hanzi},
  booktitle={ICASSP 2024-2024 IEEE International Conference on Acoustics, Speech and Signal Processing (ICASSP)},
  pages={6310--6314},
  year={2024},
  organization={IEEE}
}

@inproceedings{li2025lamot,
  title={Lamot: Language-guided multi-object tracking},
  author={Li, Yunhao and Liu, Xiaoqiong and Liu, Luke and Fan, Heng and Zhang, Libo},
  booktitle={2025 IEEE International Conference on Robotics and Automation (ICRA)},
  pages={6816--6822},
  year={2025},
  organization={IEEE}
}

@article{huang2024tell,
  title={Tell Me What to Track: Infusing Robust Language Guidance for Enhanced Referring Multi-Object Tracking},
  author={Huang, Wenjun and Ni, Yang and Chen, Hanning and He, Yirui and Bryant, Ian and Liu, Yezi and Imani, Mohsen},
  journal={arXiv preprint arXiv:2412.12561},
  year={2024}
}

@article{li2025just,
  title={Just Functioning as a Hook for Two-Stage Referring Multi-Object Tracking},
  author={Li, Weize and Du, Yunhao and Yin, Qixiang and Zhao, Zhicheng and Su, Fei and Liu, Daqi},
  journal={arXiv preprint arXiv:2503.07516},
  year={2025}
}

@article{lv2025vision,
  title={Vision-Motion-Reference Alignment for Referring Multi-Object Tracking via Multi-Modal Large Language Models},
  author={Lv, Weiyi and Zhang, Ning and Sun, Hanyang and Jiang, Haoran and Zhao, Kai and Xiao, Jing and Zeng, Dan},
  journal={arXiv preprint arXiv:2511.17681},
  year={2025}
}

@inproceedings{liu2024grounding,
  title={Grounding dino: Marrying dino with grounded pre-training for open-set object detection},
  author={Liu, Shilong and Zeng, Zhaoyang and Ren, Tianhe and Li, Feng and Zhang, Hao and Yang, Jie and Jiang, Qing and Li, Chunyuan and Yang, Jianwei and Su, Hang and others},
  booktitle={European conference on computer vision},
  pages={38--55},
  year={2024},
  organization={Springer}
}

@inproceedings{liu2023gres,
  title={Gres: Generalized referring expression segmentation},
  author={Liu, Chang and Ding, Henghui and Jiang, Xudong},
  booktitle={Proceedings of the IEEE/CVF conference on computer vision and pattern recognition},
  pages={23592--23601},
  year={2023}
}

@inproceedings{zhang2022bytetrack,
  title={Bytetrack: Multi-object tracking by associating every detection box},
  author={Zhang, Yifu and Sun, Peize and Jiang, Yi and Yu, Dongdong and Weng, Fucheng and Yuan, Zehuan and Luo, Ping and Liu, Wenyu and Wang, Xinggang},
  booktitle={European conference on computer vision},
  pages={1--21},
  year={2022},
  organization={Springer}
}

@article{zhu2020deformable,
  title={Deformable detr: Deformable transformers for end-to-end object detection},
  author={Zhu, Xizhou and Su, Weijie and Lu, Lewei and Li, Bin and Wang, Xiaogang and Dai, Jifeng},
  journal={arXiv preprint arXiv:2010.04159},
  year={2020}
}

@inproceedings{radford2021learning,
  title={Learning transferable visual models from natural language supervision},
  author={Radford, Alec and Kim, Jong Wook and Hallacy, Chris and Ramesh, Aditya and Goh, Gabriel and Agarwal, Sandhini and Sastry, Girish and Askell, Amanda and Mishkin, Pamela and Clark, Jack and others},
  booktitle={International conference on machine learning},
  pages={8748--8763},
  year={2021},
  organization={PmLR}
}

@article{luiten2021hota,
  title={Hota: A higher order metric for evaluating multi-object tracking},
  author={Luiten, Jonathon and Osep, Aljosa and Dendorfer, Patrick and Torr, Philip and Geiger, Andreas and Leal-Taix{\'e}, Laura and Leibe, Bastian},
  journal={International journal of computer vision},
  volume={129},
  number={2},
  pages={548--578},
  year={2021},
  publisher={Springer}
}

@article{hu2026curriculum,
  title={Curriculum Adaptation for One-Stream RGB--T Tracking},
  author={Hu, Xiantao and Zeng, Fansheng and Zhong, Bineng and Tang, Zhangyong and Fang, Wenxuan and Li, Jun and Tai, Ying and Yang, Jian},
  journal={Pattern Recognition},
  pages={113494},
  year={2026},
  publisher={Elsevier}
}

@inproceedings{hu2025exploiting,
  title={Exploiting multimodal spatial-temporal patterns for video object tracking},
  author={Hu, Xiantao and Tai, Ying and Zhao, Xu and Zhao, Chen and Zhang, Zhenyu and Li, Jun and Zhong, Bineng and Yang, Jian},
  booktitle={Proceedings of the AAAI Conference on Artificial Intelligence},
  volume={39},
  number={4},
  pages={3581--3589},
  year={2025}
}

@article{hu2025adaptive,
  title={Adaptive perception for unified visual multi-modal object tracking},
  author={Hu, Xiantao and Zhong, Bineng and Liang, Qihua and Shi, Liangtao and Mo, Zhiyi and Tai, Ying and Yang, Jian},
  journal={IEEE Transactions on Artificial Intelligence},
  year={2025},
  publisher={IEEE}
}

@article{hu2024toward,
  title={Toward modalities correlation for RGB-T tracking},
  author={Hu, Xiantao and Zhong, Bineng and Liang, Qihua and Zhang, Shengping and Li, Ning and Li, Xianxian},
  journal={IEEE Transactions on Circuits and Systems for Video Technology},
  volume={34},
  number={10},
  pages={9102--9111},
  year={2024},
  publisher={IEEE}
}

\clearpage
\appendix

\renewcommand{\thefigure}{A\arabic{figure}}
\renewcommand{\thetable}{A\arabic{table}}
\renewcommand{\theequation}{A\arabic{equation}}

\setcounter{figure}{0}
\setcounter{table}{0}
\setcounter{equation}{0}

\section*{Supplementary Material}

\section{Implementation Details}
This section provides a comprehensive description of the implementation details for the COAL framework. We elaborate on the data preprocessing, network architecture, training strategy, and inference procedure employed throughout the experiments.

\subsection{Data Preprocessing} 

We leverage foundation models to construct a knowledge-augmented observation space, thereby regularizing domain-specific learning with external priors. 

\begin{figure}[h]
    \centering
    \includegraphics[width=0.9\linewidth]{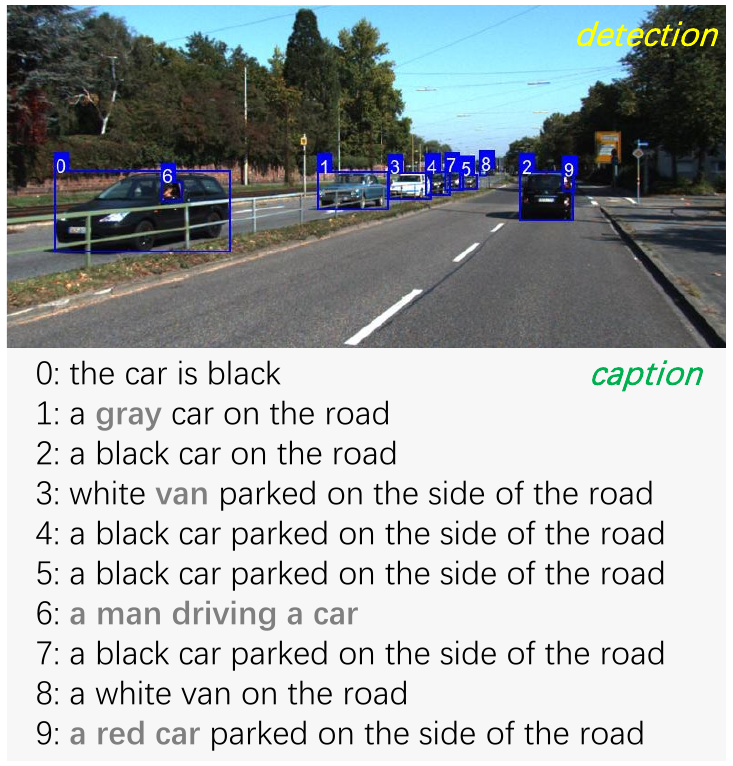}
    \vspace{-5pt}
    \caption{
    \textbf{Visualization of VLM Perception Priors.} We utilize DINO-X to extract dense observational priors, yielding comprehensive object detections and their corresponding descriptive captions. While offering rich semantic details, these raw outputs exhibit inherent noise, such as out-of-scope detections (e.g., \#6) and attribute hallucinations (e.g., \#1, \#3, \#9). This necessitates the semantic filtering performed by our Hierarchical Multi-Stream Integration (HMSI) module to distill robust discriminative cues.
    }
    \label{fig:demo_vlm}
    \vspace{-5pt}
\end{figure}

\textbf{VLM Perception Priors.} we deploy DINO-X \cite{ren2024dino} as the VLM to perform dense scene parsing, simultaneously yielding object localization and explicit attribute descriptions. As visualized in Fig.~\ref{fig:demo_vlm}, DINO-X exhibits exceptional perceptual granularity, successfully capturing a comprehensive array of instances. However, raw outputs are susceptible to intrinsic open-world noise. Visually, the detector might create clutter by identifying objects outside the RMOT scope, such as the driver in Detection \#6, which constitutes semantic noise rather than a trackable instance. Linguistically, the generated captions might suffer from hallucinations and attribute misalignment, exemplified by the inaccurate color description in Caption \#1 ("gray"), or categorical errors in Caption \#3 ("van") and Caption \#9 ("red car"). Therefore, rather than naively ingesting these raw outputs, we employ the Hierarchical Multi-Stream Integration (HMSI) architecture to perform semantic filtration and refinement. This mechanism selectively distills discriminative cues from the noisy priors, ensuring that only robust, query-aligned features contribute to the final object representation.

\begin{figure}[t]
    \centering
    \includegraphics[width=0.95\linewidth]{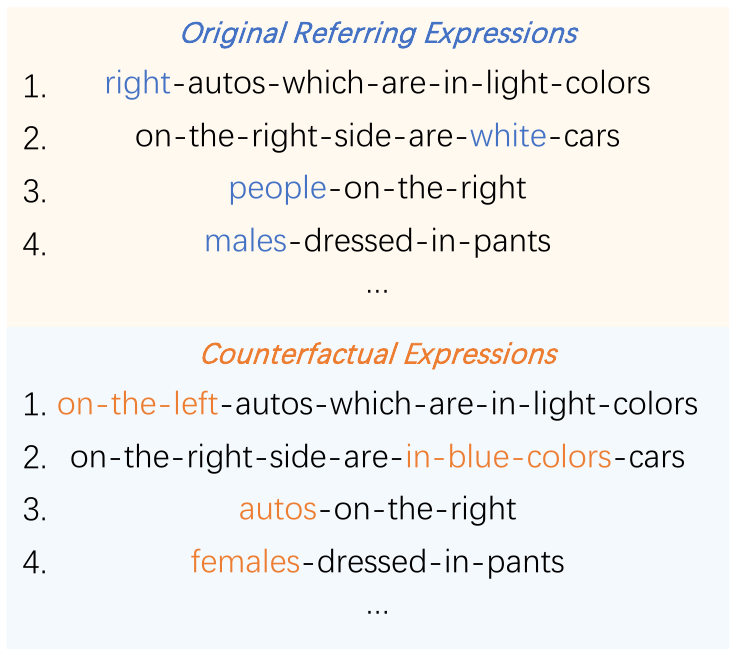}
    \vspace{-5pt}
    \caption{
    \textbf{Illustration of LLM Counterfactual Generation.} The LLM synthesizes hard negative samples by randomly replacing an attribute fragment (blue) with a context-aware counterfactual variant (orange). This stochastic perturbation generates high-interference distractors, supervising the model to acquire a comprehensive compositional understanding and prevent feature collapse caused by shortcut learning.
    }
    \label{fig:demo_llm}
    \vspace{-5pt}
\end{figure}

\textbf{LLM Counterfactual Generation.} We employ Qwen3-Max \cite{yang2025qwen3} to synthesize counterfactual expressions. For each referring expression, the LLM parses the constituent attributes and generates hard negatives by stochastically perturbing a single attribute into a plausible contrasting value. As illustrated in Fig.~\ref{fig:demo_llm}, the model modifies specific semantic components (blue), such as color, location, or object category, into contextually coherent counterfactual variants (orange). This process generates expressions that retain the structural complexity of the original query while enforcing a strict semantic divergence. By introducing these high-interference distractors, we compel the model to refine its fine-grained discrimination boundaries.

\subsection{Network Architecture} 
We utilize a frozen CLIP (ResNet-50) \cite{radford2021learning} as the backbone for multi-modal feature extraction.
On the visual side, we construct a four-layer Feature Pyramid Network (FPN) with $d=256$ channels using multi-scale features derived from the encoder. The deformable sampling \cite{zhu2020deformable} follows the standard configuration (4 levels, 8 heads) to extract geometry-aware features.
On the linguistic side, we extract both word-level and sentence-level embeddings for referring expressions, counterfactual queries, and VLM captions.
Across the Pixel-Word Contextualization and Hierarchical Semantic Refinement, all cross-attention blocks maintain a unified hidden dimension of $d=256$ with 8 heads to ensure consistent feature interaction.

\begin{figure}[t]
    \centering
    \includegraphics[width=1.0\linewidth]{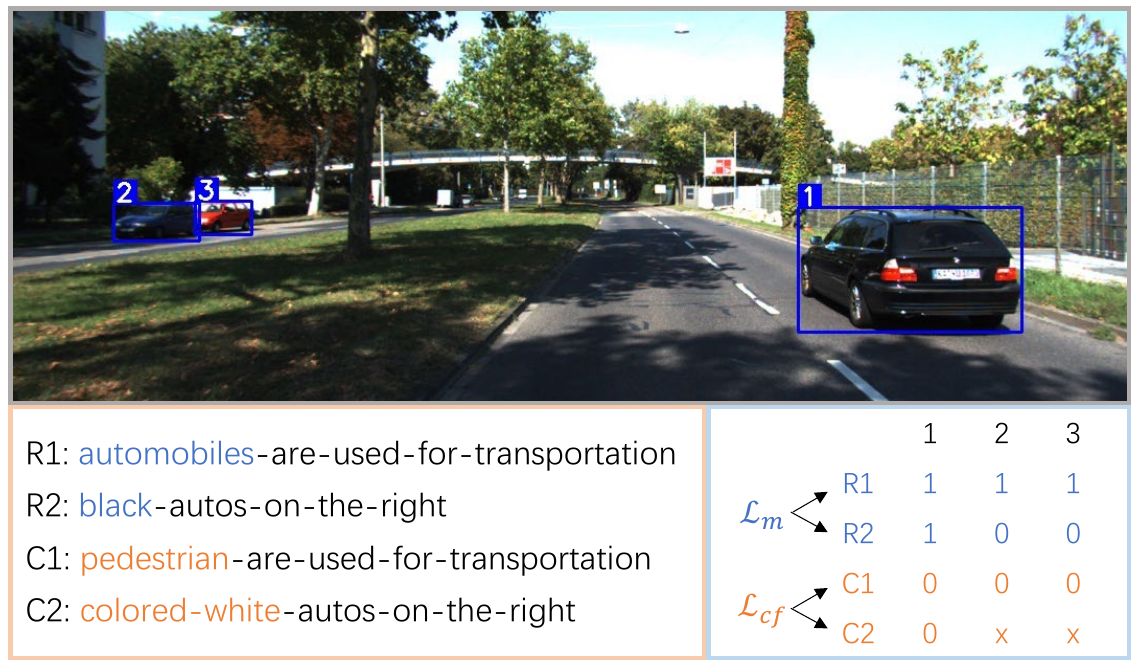}
    \vspace{-5pt}
    \caption{
    \textbf{Illustration of the 1-Image-2$N$-Queries Training Strategy ($N=2$).} For a given image with objects $\{1, 2, 3\}$, we construct a query batch containing positive expressions ($R_1, R_2$) and their counterfactual negatives ($C_1, C_2$). The label assignment matrix illustrates the dual supervision: $\mathcal{L}_m$ applies standard supervision across all objects, whereas $\mathcal{L}_{cf}$ selectively targets the original positive instance by forcing its label to 0, while masking other objects with label "x" to prevent false penalties. 
    }
    \label{fig:demo_training}
    \vspace{-5pt}
\end{figure}

\subsection{Training Strategy} 
We propose a 1-Image-2$N$-Queries strategy for efficient training, as shown in Fig.~\ref{fig:demo_training}. In each iteration, a single image is paired with a batch of $2N$ queries: $N$ annotated positive expressions (e.g., $R_1, R_2$) and their corresponding $N$ counterfactual negatives (e.g., $C_1, C_2$). Crucially, the label assignment operates under two distinct protocols:
(1) The primary loss ($\mathcal{L}_m$) is computed for positive expressions ($R_1, R_2$) across all objects. Labels are assigned via Hungarian matching based on Intersection over Union (IoU) between VLM-detected boxes and ground-truth annotations.
(2) The counterfactual loss ($\mathcal{L}_{cf}$) is computed for counterfactual queries ($C_1, C_2$). We enforce a strict penalty merely on the original target object by inverting the label from 1 to 0, while masking all other candidates as label "x". This masking is crucial to prevent false penalties, as a randomly perturbed query could coincidentally describe another valid object in the scene. 

This strategy maximizes knowledge density while minimizing redundant visual computation.
Across all experiments, we set $N=10$. The model is trained for 30 epochs on 4 NVIDIA L40 GPUs using the AdamW optimizer. The learning rate is fixed at $10^{-4}$ throughout training. To ensure reproducibility, we set the random seed to 42.

\subsection{Inference Procedure} 
We adopt the standard Tracking-by-Detection (TBD) paradigm, employing COAL as an enhanced observation model to maximize the semantic discriminability of detections. By computing precise similarity scores between candidates and referring expressions, COAL provides high-quality grounding foundations for the subsequent association. The inference pipeline of COAL mirrors the training phase, simply excluding the counterfactual supervision branch. For trajectory linking, we integrate the ByteTrack \cite{zhang2022bytetrack} algorithm, which associates objects based on these predicted semantic scores and bounding box IoU without requiring additional tracker training. Following the ByteTrack protocol, we maintain the detection score thresholds of $\tau_{high}=0.4$, $\tau_{low}=0.1$, and the tracking score threshold $\epsilon=0.4$ across all experiments.
The source code and model weights will be made publicly available for reproducibility.

\end{document}